\documentclass[letterpaper, 10 pt, conference]{ieeeconf}  
\IEEEoverridecommandlockouts

\IEEEoverridecommandlockouts                              

\usepackage{graphics} 
\usepackage{graphicx}
\usepackage{amsmath} 
\usepackage{amssymb}  
\usepackage{booktabs}
\usepackage{multirow}
\usepackage{algorithm}
\usepackage{algorithmic}
\usepackage{url}
\usepackage{xcolor}

\title{\LARGE \bf
Enhancing LLM Problem Solving via Tutor–Student Multi-Agent Interaction}

\author{Nurullah Eymen Özdemir$^{1}$ and Erhan Oztop$^{1,2}$
\thanks{This work is under review for conference appearance.}
\thanks{$^{1}$Nurullah Eymen Özdemir and Erhan Oztop is with Ozyegin University, Istanbul, Turkiye
        {\tt\small eymen.ozdemir@ozu.edu.tr, erhan.oztop@ozyegin.edu.tr}}%
\thanks{$^{2}$Erhan Oztop is also affiliated affiliated with Symbiotic
Intelligent Systems Research Center, Institute for Open and Transdisciplinary, Research Initiatives, The University of Osaka, Japan.
        {\tt\small erhan.oztop@otri.osaka-u.ac.jp}}%
}

\begin{document}

\maketitle
\thispagestyle{empty}
\pagestyle{empty}

\begin{abstract}

 Human cognitive development is shaped not only by individual effort but by structured social interaction, where role-based exchanges such as those between a tutor and a learner, enable solutions that neither could achieve alone. Inspired by these developmental principles, we ask the question whether 
 a tutor-student multi-agent system can create a synergistic effect by pushing Large Language Model (LLM)  beyond what it can do within existing frameworks. To test the idea, we adopt autonomous coding problem domain where two agents instantiated from the same LLM assigned asymmetric roles: a student agent generates and iteratively refines solutions, while a tutor agent provides structured evaluative feedback without access to ground-truth answers. In our proposed framework (PETITE), we aim to extract better problem-solving performance from one model by structuring its interaction through complementary roles, rather than relying on stronger supervisory models or heterogeneous ensembles.
 Our model is evaluated on the APPS coding benchmark against state-of-the-art approaches of Self-Consistency, Self-Refine, Multi-Agent Debate, and Multi-Agent Review. The results show that our model achieves similar or higher accuracy while consuming significantly fewer tokens. These results suggest that developmentally grounded role-differentiated interaction structures provide a principled and resource-efficient paradigm for enhancing LLM problem-solving through structured peer-like interactions.

 
\end{abstract}

\section{INTRODUCTION}

\begin{figure*}[t]
\centering
\includegraphics[width=0.9\textwidth]{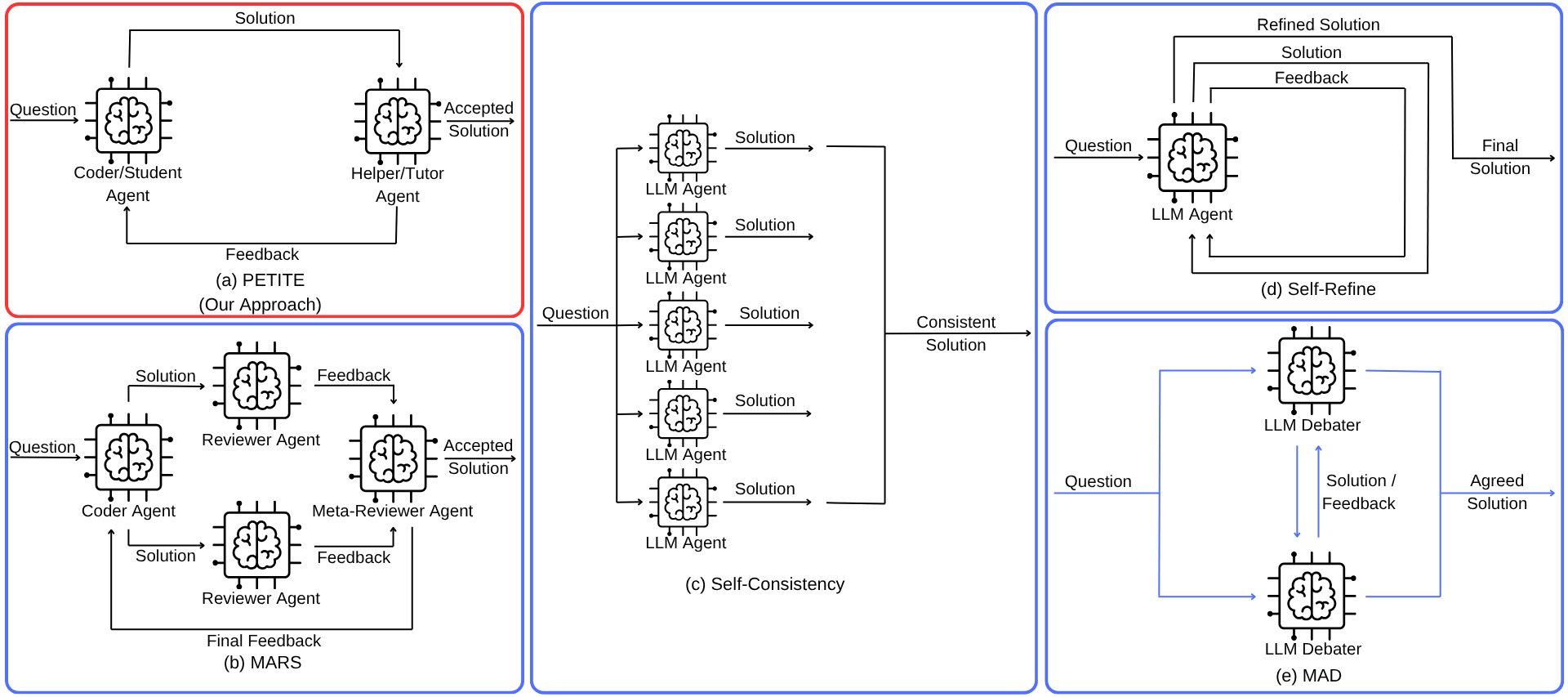}
\caption{ The proposed PETITE framework and considered baseline architectures. (a) In PETITE a student (coder) generates solutions and a tutor (helper) provides feedback in an iterative loop, producing the final accepted solution. (b) MARS introduces structured roles, including coder, reviewer, and meta-reviewer agents, to perform hierarchical evaluation and refinement. (c) Self-Consistency samples multiple independent solutions and selects the most consistent one. (d) Self-Refine employs a single LLM that iteratively generates, critiques, and refines its own solution. (e) Multi-Agent Debate (MAD) enables multiple agents to iteratively exchange solutions and feedback until reaching consensus.}
\label{fig:algorithms_schema}
\end{figure*}

A central principle of developmental science is that cognitive growth does not occur in isolation, but emerges through structured social interaction \cite{vygotsky1978mind, piaget1952origins}. The concept of the zone of proximal development (ZPD) \cite{vygotsky1978mind} captures this idea by positing that learners can exceed their independent capabilities when supported by appropriate guidance \cite{wood1976tutoring}. While the concept is typically framed in terms of a more knowledgeable other, peer tutoring research shows that structured role differentiation among similarly skilled learners can also produce significant learning gains, even in the absence of capability asymmetry \cite{topping2005peer}. This effect can be understood through role specialization: adopting an evaluative role promotes analytical reasoning, which in turn supports more effective solution generation. More generally, one-on-one tutoring is widely recognized as one of the most effective instructional formats \cite{bloom1984two}, largely due to the availability of continuous, structured feedback during interaction, which improves learning regardless of whether it is provided by a more knowledgeable tutor or a peer \cite{black1998formative}. Taken together, these studies suggest that guided interaction, role differentiation, and continuous evaluation are key concepts of human learning and development.

These developmental principles have inspired a growing body of work in artificial intelligence and machine learning. Developmental robotics \cite{cangelosi2015developmental, weng2001autonomous} explores how artificial agents can acquire human-like learning behaviors through sensorimotor and social interaction. Similarly, cognitive apprenticeship frameworks [7] have shaped the design of AI tutoring systems that make expert reasoning explicit to support learning. 
Despite these connections, recent advances in Large Language Model (LLM) multi-agent systems have yet to incorporate principles from developmental learning, leaving the potential benefits of structuring interaction design based on human cognitive development largely unexplored.

To investigate whether developmentally inspired interaction structures can improve LLM performance, we instantiate the approach in the context of autonomous code generation. We design a multi-agent setup in which two agents assume complementary roles: a student that generates and iteratively refines solutions, and a tutor that evaluates and provides guidance. This setup enables controlled analysis of how role differentiation and iterative feedback influence performance within an LLM-based system.

While LLMs have demonstrated strong capabilities in automated code generation \cite{chen2021codex, li2022alphacode}, they still struggle to achieve consistently high accuracy on complex algorithmic problems, especially at smaller model scales. Current approaches to improving LLM performance include enhanced prompting strategies such as Chain-of-Thought reasoning \cite{wei2022cot}, sampling-based methods like Self-Consistency \cite{wang2023selfconsistency}, and iterative refinement frameworks such as Self-Refine \cite{madaan2023selfrefine}, as well as multi-agent systems including Multi-Agent Debate \cite{liang2023mad, du2023debate} and MARS \cite{du2023mars}. While these methods have shown gains on several benchmarks, our experiments indicate that they can be fragile, often failing to improve performance and in some cases degrading accuracy in code generation tasks, where the solution space is open-ended and correctness admits many valid implementations.
Among existing approaches, many employ symmetric or consensus-driven interaction patterns \cite{liang2023mad, du2023debate} that do not explicitly capture interaction dynamics inspired by human learning. MARS \cite{du2023mars} introduces asymmetric roles but still relies on agreement among parallel reviewers, incurring substantial token overhead despite incorporating an early stopping mechanism. To our knowledge, these methods do not adopt role-differentiated, serial scaffolding structures commonly studied in human learning \cite{bloom1984two, wood1976tutoring, topping2005peer}.

Beyond adherence to developmental principles, our approach is also motivated by a common observation in human problem-solving: when writing code, individuals often overlook errors in their own solutions, yet readily identify similar mistakes when evaluating others’ work. This asymmetry between generative and evaluative processes is well documented in peer tutoring research \cite{topping2005peer}. Evaluation engages analytical attention to correctness, edge cases and logical consistency, which is suppressed during the generative process of constructing a solution. We hypothesize that a similar asymmetry applies to LLMs: a single model, when prompted to evaluate code rather than generate it, may detect errors it would miss. If so, structuring the interaction between a generator and an evaluator roles could yield systematic performance without increasing model capacity.

We introduce our PETITE (\textbf{Pe}er \textbf{T}utoring \textbf{I}nspired \textbf{T}oken-\textbf{E}fficient) framework and instantiate it for code generation, a multi-agent system that balances performance gains with token efficiency. PETITE instantiates two functionally differentiated roles: \emph{Student/Coder Agent} is responsible for generating and refining code solutions based on the problem description and received feedback while \emph{Tutor/Helper Agent} evaluates the student's solutions, identifies errors, logical gaps and edge cases, and provides structured feedback to guide improvements. A key innovation of PETITE is its serial usage of helper agent with asymmetric role assignment. Unlike traditional iterative approaches that execute a fixed number of refinement rounds, PETITE terminates when the tutor agent determines that the solution is correct. This adaptive termination naturally allocates more computational resources to harder problems while efficiently resolving simpler ones and reduces unnecessary token consumption, mirroring the differential pacing observed in human cognitive development.

Our contributions are as follows:
\begin{itemize}
\item We propose a peer tutoring inspired LLM framework (PETITE), instantiated in code generation domain, a multi-agent system that uses structured role-based interaction for iterative code refinement.
\item We show that asymmetric tutor–student role separation, motivated by peer tutoring and scaffolding theory, provides an effective alternative to symmetric debate or parallel review architectures, even with identical agent capabilities.
\item We introduce an early-stopping mechanism based on tutor evaluation without ground truth, enabling adaptive control of computational resources.
\item We evaluate PETITE on the APPS benchmark, demonstrating competitive accuracy with significantly lower token usage than existing methods.
\end{itemize}

\section{RELATED WORK}

\subsection{Prompting Strategies for Code Generation}

Chain-of-Thought (CoT) prompting \cite{wei2022cot} has proven effective for complex reasoning tasks by encouraging models to generate intermediate reasoning steps before producing the final answer. For code generation, CoT prompting guides models to decompose problems, identify constraints, and develop algorithmic strategies before implementation. While CoT improves accuracy on reasoning-intensive problems, it operates in a single inference pass and cannot correct errors post-generation.

\subsection{Self-Consistency and Sampling Methods}

Self-Consistency \cite{wang2023selfconsistency} addresses the limitation of single-pass generation by sampling multiple reasoning paths and selecting the most consistent answer through majority voting. For code generation, this involves generating multiple independent solutions and selecting based on output's success rate agreement. While effective, Self-Consistency's computational cost scales linearly with the number of generation repetitions, making it expensive for deployment scenarios.

\subsection{Iterative Refinement Approaches}

Self-Refine \cite{madaan2023selfrefine} proposes an iterative framework in which a model generates an initial solution, critiques its own output, and refines it based on the feedback. The same model alternates between solver and critic roles, enabling improvement without additional training.
Related work explores structured self-evaluation mechanisms. Self-Reflection \cite{renze2024selfreflection} studies the impact of reflective reasoning steps on problem-solving performance, encouraging models to reconsider prior reasoning before finalizing answers. Self-Verification \cite{weng2023selfverification} further separates solution generation and verification, introducing explicit consistency or logic checks before refinement.
Among these methods, Self-Refine is the most appropriate baseline for comparison with PETITE, as both rely on iterative feedback-driven improvement.

\subsection{Multi-Agent Systems}

Multi-Agent Debate (MAD) \cite{liang2023mad, du2023debate} employs multiple LLM instances that engage in structured debates to reach consensus on complex problems. Each agent proposes solutions and critiques others' proposals until agreement is reached. While effective for reasoning tasks, MAD can be token-intensive due to extended debate rounds.

MARS (Multi-Agent Review System) \cite{du2023mars} introduces a hierarchical review structure with multiple reviewers and a meta-reviewer that synthesizes feedback. The solver then refines solutions based on consolidated critiques. This structured approach improves feedback quality but increases token consumption through multiple review phases and parallel connection of the reviewers.

\subsection{Position of Our Work}

In addition to increasing the success rate of solutions, PETITE distinguishes itself from existing approaches through its focus on token efficiency without sacrificing accuracy. While MAD and MARS employ symmetric agent interactions or consensus based structures, PETITE adopts an asymmetric tutor-student paradigm that naturally models the teaching-learning dynamic. The tutor provides targeted feedback focused on correctness, logic gaps and edge cases, while the student implements improvements.

\section{METHODOLOGY}

\subsection{Problem Formulation}
  
\begin{figure}[t]
\centering
\includegraphics[width=\columnwidth,height=0.7\textheight,keepaspectratio]{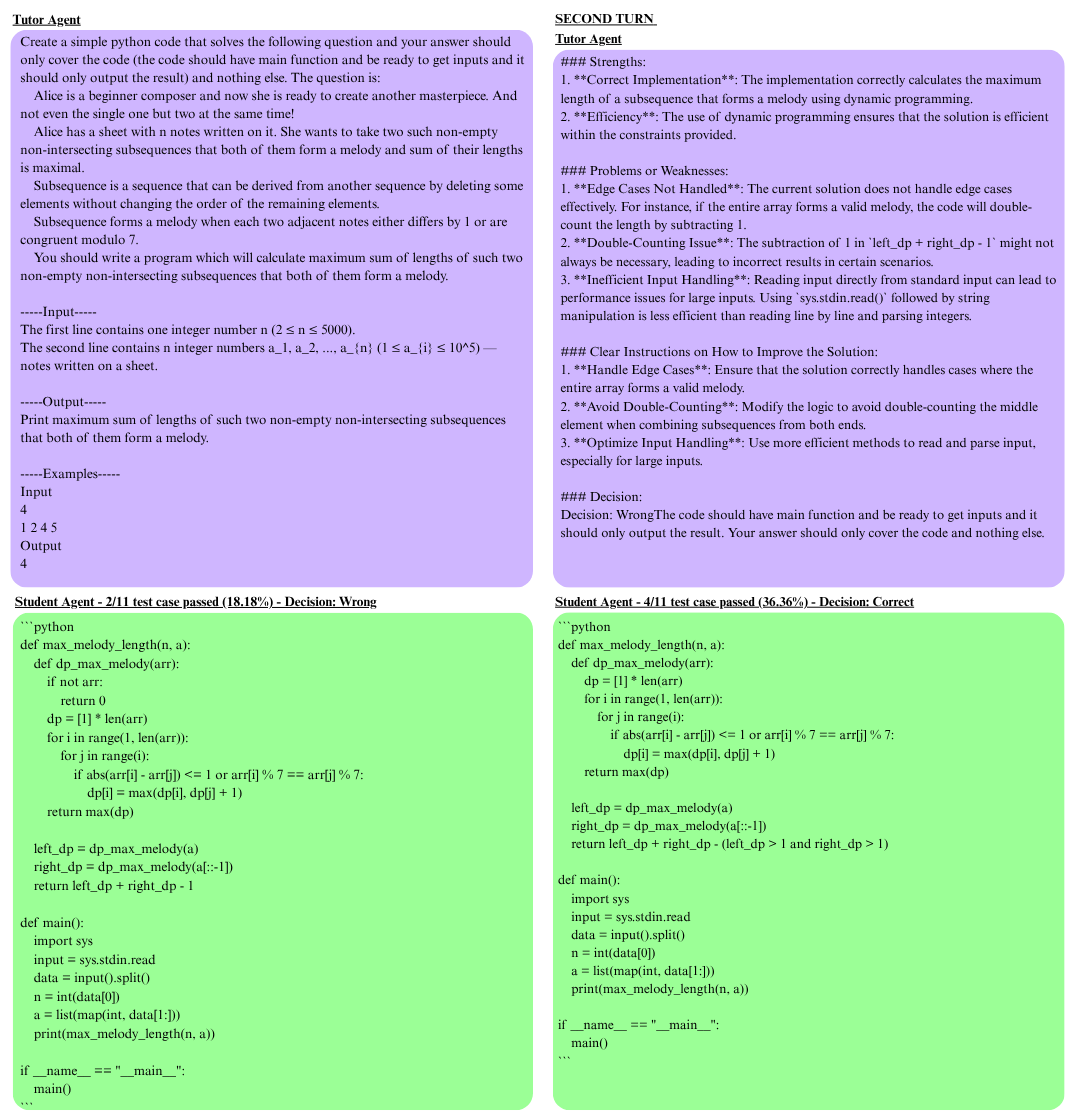}
\caption{A successful refinement interaction by our model is depicted. After the initial response of the student agent, the tutor agent evaluated the code generated and provided feedback to improve the student's solution. At the second iteration, tutor found the code regenerated by the student sufficient and labeled the solution as "Correct".} 
\label{fig:refinement}
\vspace{-4mm}
\end{figure}

A programming problem consists of a natural language specification and a set of input–output test cases that define the expected behavior of a correct solution. The goal is to generate a program that satisfies all provided test cases. We conduct experiments on the APPS benchmark \cite{hendrycks2021apps}, where each problem includes multiple input–output pairs. Fig.~\ref{fig:refinement} shows an example refinement process with the problem instance. Solution correctness is determined by the fraction of test cases passed, as specified by the dataset. We also report token efficiency, measured as the total number of tokens consumed across all inference calls during the solution process. 

\subsection{LLM Prompts for PETITE based Code Generation}

Our algorithm consists of two agents instantiated from the same base LLM but with distinct system prompts defining their roles:

\emph{Student Agent System Prompt:}
\begin{quote} \small
``You are a coder. When user asks you questions try to solve them perfectly. Regarding the user's feedbacks update your answer. IMPORTANT: You are not allowed to write anything else than code. Do not write manual test cases. Ensure the final code satisfies all constraints. Provide only the final code in a single code block.''
\end{quote}

\emph{Tutor Agent System Prompt:}
\begin{quote} \small
``You are a coding assistant and expert in coding. Your job is to evaluate the solver's answer and find ALL issues. Identify all errors, logical errors, syntax errors, gaps, infinite loops, unclear reasoning, edge cases, range issues. Provide constructive and actionable feedback. Your output must include: 1) List of strengths (if any), 2) List of problems or weaknesses, 3) Clear instructions on how to improve the solution, 4) At the end, state `Decision: Wrong' if the code is wrong, `Decision: Correct' if the code is correct.''
\end{quote}

\subsection{Early Stopping Mechanism}

The early stopping mechanism is triggered when the tutor’s feedback includes the phrase “Decision: Correct,” indicating that the current solution is considered satisfactory. This decision is made solely based on the tutor agent’s evaluation, without access to any ground-truth solution. This mechanism improves efficiency by terminating the refinement process as soon as a correct solution is identified, thereby reducing unnecessary token usage. It also helps prevent performance degradation, since continued refinement of an already correct solution may introduce new errors. In addition, it enables adaptive allocation of computational resources, as more iterations are naturally spent on harder problems while simpler ones are resolved earlier.

\subsection{Iterative Refinement Process}

The PETITE process proceeds as shown in Alg. 1, the procedure begins by initializing two separate conversation contexts for the student and tutor agents, each with their respective system prompts. The given problem description is formatted and provided to both agents, ensuring that they start from the same task context. At each iteration, the student agent generates a candidate solution based on its current conversation history. This solution is then passed to the tutor agent, which evaluates it and produces structured feedback. The tutor’s feedback includes both qualitative assessment and a decision signal indicating whether the solution is correct. If the tutor’s feedback contains the phrase “Decision: Correct,” the process is terminated early, and the current solution is returned. Importantly, this decision is made solely based on the tutor agent’s internal evaluation, without access to ground-truth answers. If the solution is not accepted, the tutor’s feedback is appended to both agents’ contexts. The student then uses this feedback to refine its previous attempt in the next iteration. This iterative interaction continues until a correct solution is identified or the maximum number of iterations is reached.

\begin{algorithm}[h]
\footnotesize
\caption{PETITE Tutor-Student Algorithm}
\begin{algorithmic}[1]
\setcounter{ALC@line}{-1}
\STATE ($S_t$, $F_t$ be "Solution" and "Feedback" generated at step $t$)
\STATE \textbf{Input:} Problem description $P$, max iterations $T$
\STATE \textbf{Output:} Final solution $S$
\STATE Initialize student conversation $C_S$ with system prompt
\STATE Initialize tutor conversation $C_T$ with system prompt
\STATE $prompt \leftarrow$ FormatProblem($P$)
\STATE Append $prompt$ to $C_S$ as user message
\STATE Append $prompt$ to $C_T$ as assistant message
\FOR{$t = 1$ to $T$}
    \IF{$t > 1$}
        \STATE Append $S_{t-1}$ to $C_T$ as user message
        \STATE $F_t \leftarrow$ TutorAgent.Generate($C_T$) \COMMENT{Feedback}
        \IF{``Decision: Correct'' in $F_t$}
            \STATE \textbf{break} \COMMENT{Early stopping}
        \ENDIF
        \STATE Append $F_t$ to $C_S$ as user message
        \STATE Append $F_t$ to $C_T$ as assistant message
    \ENDIF
    \STATE $S_t \leftarrow$ StudentAgent.Generate($C_S$)
    \STATE Append $S_t$ to $C_S$ as assistant message
\ENDFOR
\STATE \textbf{return} $S_t$
\end{algorithmic}
\end{algorithm}
\subsection{Token Tracking}
To assess efficiency, we track token usage for both agents, including input and output tokens at each iteration and cumulative totals over the course of interaction. Token counts are recorded separately for the student and tutor agents to analyze computational cost and compare with baseline methods.
%

\section{EXPERIMENTAL SETUP}

\begin{figure}[t]
\centering
\includegraphics[width=\columnwidth,height=0.5\textheight]{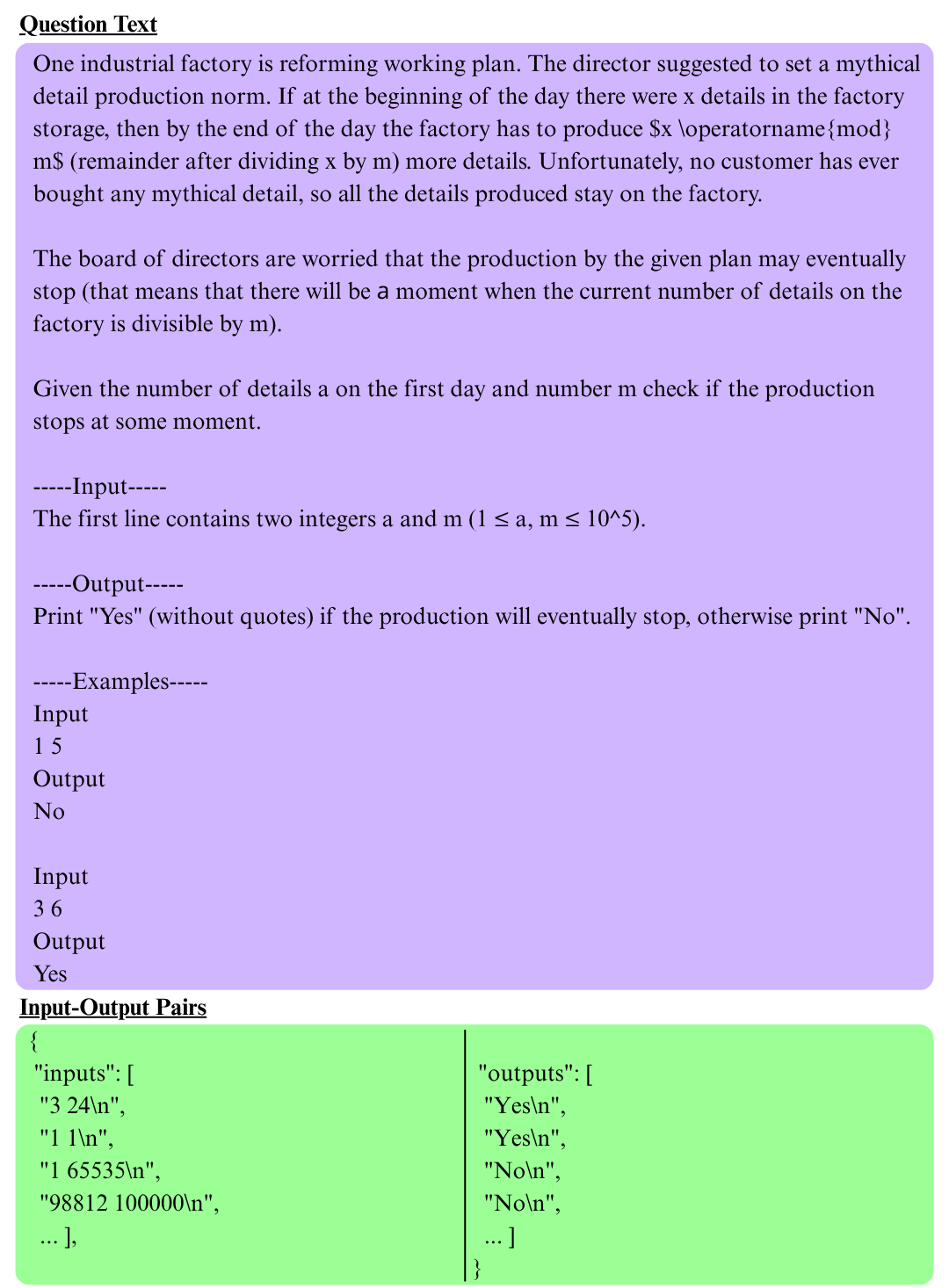}
\caption{Example of an "Interview" level problem from META APP database. Each problem entry contains the question body, associated metadata, input–output test pairs for evaluation (truncated for the sake of space).} 
\label{fig:database_single}
\end{figure}

\subsection{Benchmark Dataset}

We evaluate on a subset of 100 problems randomly sampled from the APPS benchmark \cite{hendrycks2021apps}, which consists of coding problems sourced from competitive programming platforms including META's coding challenges. The problems span three difficulty levels determined by the META: \emph{Introductory} (14\%) level problems designed as basic programming concepts while \emph{Interview} (62\%) level represents technical interview difficulty and \emph{Competition} (24\%) stands for competitive programming level. Each problem includes a natural language description, input/output specifications, and multiple test cases for evaluation.Fig.~\ref{fig:database_single} shows an example problem instance with truncated input-output test pairs.

\subsection{Base LLM Model Details}
All experiments use Qwen2.5-Coder-7B-Instruct [21] as the base model, deployed with 4-bit quantization (NF4) to enable efficient inference on consumer-grade hardware (RTX 4060 with 8GB VRAM). The maximum context length is set to 4096 tokens, and the generation is limited to 2048 new tokens per request. We use stochastic decoding with sampling enabled (do\_sample=True) and a temperature in the range of 0.7 to 0.8.

\subsection{Baseline Methods}

We compare our framework, PETITE against four baseline approaches. (See Fig.~\ref{fig:algorithms_schema}.) In brief, the baselines' operation logic is as follows:

\noindent \emph{Self-Consistency (SC)}: Generation of 10 independent solutions with mode-based selection. We report both 5-sample and 10-sample consistency.

\noindent \emph{Self-Refine (SR)}: Three-phase approach with Solver, Critic, and Refiner roles operating sequentially for 2 iterations.

\noindent \emph{Multi-Agent Debate (MAD)}: Two debating agents generate solutions and iteratively refine based on each other's proposals until consensus (matching success rates) or maximum 10 rounds.

\noindent \emph{MARS}: Multi-agent review system with one solver, two parallel reviewers, and a meta-reviewer that synthesizes feedback.

\subsection{Evaluation Metrics}
We evaluate performance using two complementary metrics. \textit{Success Rate} is the percentage of test cases passed per problem, averaged across the benchmark. \textit{Improvement} is defined as the difference between the success rate of the final and initial solutions. As initial solutions vary due to high sampling temperatures, \textit{Improvement} better reflects the effect of the refinement process. To account for computational cost, we measure \textit{Token Consumption} as the total number of input and output tokens per problem, averaged over the dataset. We also define an \textit{Efficiency Ratio} as the success rate divided by token consumption to quantify performance relative to computational cost.


\section{RESULTS}
In this section, we compare PETITE with the baseline methods on the APPS benchmark. All methods use the same base LLM and generation settings (Section IV) to ensure a fair comparison. We report effectiveness (success rate and improvement) and efficiency (token consumption and efficiency ratio), with results presented across overall performance, difficulty levels, and token usage.

\subsection{Overall Performance Comparison}
The experimental results show that our framework, PETITE consistently outperforms or matches the baselines, with MARS being the closest in terms of performance. Table \ref{tab:overall_efficiency} summarizes the results across all methods. All methods are evaluated under the same experimental setup, using identical model configurations and decoding parameters (see Section~IV). The numbers in parentheses indicate method-specific configurations: for Self-Consistency it implies number of repetitions; for Self-Refine and PETITE it indicates number of maximum refinement iterations. 
\begin{table}[h]
\caption{Overall Performance and Efficiency (100 Problems)}
\label{tab:overall_efficiency}
\begin{center}
\vspace{-5mm}
\begin{tabular}{lcccc}
\toprule
\textbf{Method} & \textbf{Success (\%)} & \textbf{Improve.} & \textbf{Avg Tokens} & \textbf{Efficiency} \\
\midrule
Self-C. (5)  & 27.50 & -2.88 & 5,407.48  & 5.09 \\
Self-C. (10) & 28.84 & -1.71 & 10,831.37 & 2.66 \\
Self-Refine (1)       & 29.45 $\pm$ 1.54 & -0.46 & 5,398.70  & 5.45 \\
Self-Refine (2)       & 27.59 $\pm$ 1.89 & -2.32 & 15,285.50 & 1.80 \\
MAD                   & 28.97 $\pm$ 1.96 & --  & 23,630.50 & 1.23 \\
MARS                  & 30.06 $\pm$ 2.65 & -0.82 & 7,319.99  & 4.11 \\
PETITE (1)             & 31.13 $\pm$ 2.48 & 0.34  & \textbf{2,490.90}  & \textbf{12.50} \\
PETITE (2)             & \textbf{31.62 $\pm$ 2.29} & \textbf{0.83} & 5,277.20  & 5.99 \\
PETITE (3)             & 31.24 $\pm$ 2.30 & 0.45  & 9,622.40  & 3.25 \\
\bottomrule
\end{tabular}
\end{center}
\vspace{-5mm}
\end{table}

Inspecting results in Table \ref{tab:overall_efficiency} one can note that Self-Consistency exhibits diminishing returns as the number of samples increases, with 10-sample consistency providing only a marginal improvement over 5-sample consistency while nearly doubling token usage. Similarly, increasing the refinement depth in Self-Refine (from one to two iterations) leads to performance degradation alongside a significant increase in computational cost.

Our model, PETITE, demonstrates consistent performance improvements across refinement iterations, with the two-iteration configuration achieving the highest overall success rate (31.62\%). Notably, even PETITE (1) outperforms several multi-agent baselines while consuming considerably fewer tokens. Additional iterations (PETITE (3)) increase computational cost without proportional gains, indicating that two iterations offer the most favorable balance between accuracy and efficiency. The Improvement column in the table indicates the difference between the final success rate and the initial success rate of the corresponding method, capturing the effectiveness of iterative refinement in enhancing solution quality. 

\subsection{Performance by Difficulty}
To better understand how each method performs across different levels of problem complexity, we analyze results based on difficulty categories defined in the APPS benchmark. 
\begin{table}[h]
\caption{Success Rate (\%) by Problem Difficulty}
\vspace{-4mm}
\label{tab:difficulty}
\begin{center}
\begin{tabular}{lccc}
\toprule
\textbf{Method} & \textbf{Intro} & \textbf{Interview} & \textbf{Comp} \\
\midrule
Self-Consistency (5) & 48.07 & 28.17 & 13.75 \\
Self-Consistency (10) & 49.58 & 28.94 & 16.50 \\
Self-Refine (2) & 38.50 $\pm$ 6.65 & 28.35 $\pm$ 2.27 & 19.28 $\pm$ 3.91 \\
MAD & \textbf{52.94 $\pm$ 5.62} & 27.26 $\pm$ 2.47 & 19.39 $\pm$ 3.63 \\
MARS & 49.46 $\pm$ 5.83 & 29.01 $\pm$ 2.67 & 21.45 $\pm$ 4.31 \\
PETITE (2) & 52.68 $\pm$ 5.62 & \textbf{30.77 $\pm$ 2.62} & \textbf{21.50 $\pm$ 2.87} \\
\bottomrule
\end{tabular}
\end{center}
\end{table}

Table \ref{tab:difficulty} presents the success rates of different methods across problem difficulty levels, as defined by META's APPS benchmark. The benchmark categorizes problems into Introductory, Interview, and Competition levels, with the interview set being the most populated and most representative of real-world coding scenarios.
PETITE demonstrates particularly strong performance on Introductory problems (52.68\%), which are typically well-handled by the initial student responses and therefore require minimal refinement. On the Interview set, PETITE achieves the largest improvement relative to initial responses (30.77\%), reflecting the effectiveness of the tutor-student iterative process for problems of moderate complexity, where targeted feedback can meaningfully increase solution quality. Performance on Competition problems (21.50\%) remains competitive, although the inherent difficulty of these problems limits the gains achievable even with iterative refinement. 

\subsection{Token Consumption Analysis}

In addition to accuracy, computational cost, measured via token consumption, is an important factor in evaluating LLM-based methods. To assess efficiency, we compare token consumption across methods and problem difficulty levels. Table \ref{tab:tokens} summarizes the average token usage for each method. It can be seen that PETITE, demonstrates consistent token efficiency across all difficulty levels.  Notably, on introductory problems, it consumes 3,078 tokens, substantially fewer than MARS (5,525) and MAD (15,251).

\begin{table}[h]
\caption{Average Token Consumption by Difficulty}
\vspace{-4mm}
\label{tab:tokens}
\begin{center}
\begin{tabular}{lcccc}
\toprule
\textbf{Method} & \textbf{Intro} & \textbf{Interview} & \textbf{Comp} & \textbf{Overall} \\
\midrule
Self-Consistency (5) & 4,195 & 5,482 & \textbf{5,923} & 5,407 \\ 
Self-Consistency (10) & 8,373 & 10,964 & 11,924 & 10,831 \\ 
Self-Refine (2) & 12,828 & 15,413 & 16,390 & 15,286 \\ 
MAD & 15,251 & 26,767 & 20,415 & 23,631 \\ 
MARS & 5,525 & 7,244 & 8,563 & 7,320 \\ 
PETITE (2) & \textbf{3,078} & \textbf{5,040} & 7,174 & \textbf{5,277} \\
\bottomrule
\end{tabular}
\end{center}
\end{table}

For Interview problems, the most representative real-world problem pool, PETITE uses 5,040 tokens, achieving a reduction of approximately 81\% compared to MAD (26,767) and 67\% compared to Self-Refine (15,413)  with better performance as shown in Table \ref{tab:difficulty}. For Competition problems, PETITE requires 7,174 tokens, markedly fewer than Self-Refine (16,390) and MAD (20,415) still keeping its higher efficiency and performance in this challenging problem set.


\section{DISCUSSION}


Our results highlight how structuring interaction between agents affects both performance and efficiency in code generation tasks. PETITE framework, implemented within code generation domain, 
consistently achieves strong performance while maintaining low token consumption. This behavior can be attributed to its serial interaction pattern, where a student agent generates solutions and a tutor agent evaluates them and provides targeted feedback. This separation of roles allows the model to focus on generation and evaluation as distinct processes, leading to more stable refinement across iterations.

We observe that a small number of refinement steps is sufficient to capture most of the gains. Through empirical evaluation, we found that two refinement iterations provide the best balance between accuracy and token efficiency; additional iterations (three or four) did not yield consistent improvements relative to their computational cost.
We further explored temperature settings between 0.5 and 0.9 and observed that a temperature of 0.8 produced the best performance. Lower temperatures often led to similar solution patterns and recurring mistakes, while slightly higher randomness introduced beneficial diversity that occasionally enabled improved refinements. Although resetting the tutor’s conversation history at each iteration would further reduce token usage, we retained cumulative context to preserve the conceptual integrity of the tutor–student interaction.
By adopting sequential refinement, PETITE mitigates consensus-driven drift toward suboptimal solutions, in contrast to parallel or debate-based approaches that can exhibit ill-convergence due to premature or unstable agreement. Combined with the decision-based early stopping mechanism, this design strengthens token efficiency while maintaining competitive performance relative to alternative multi-agent approaches.

The asymmetric tutor–student interaction in our framework provides a more controlled alternative. By assigning distinct responsibilities to each agent, the framework avoids the need for consensus and instead focuses on iterative improvement guided by evaluation. Feedback flows in a single direction, which simplifies the interaction and reduces the risk of circular reasoning. The tutor’s decision also provides a natural stopping point, allowing the system to adapt its computation based on problem difficulty. Reviewer-based approaches such as MARS benefits role asymmetry similar to our approach, however, the primary distinction lies in reviewer coordination. MARS utilizes multiple reviewers in parallel, whereas PETITE applies feedback in a serial manner. This parallel structure increases token consumption per case, even in their minimal configuration consisting of two reviewers and one meta-reviewer. Consequently, the use of multiple reviewers in parallel increases computational cost and may still lead to convergence toward broadly acceptable but not necessarily optimal solutions.

Despite these advantages, several limitations remain. The effectiveness of our approach depends on the reliability of the tutor agent’s judgments. If the tutor incorrectly marks a solution as correct, the process may terminate prematurely and reduce overall accuracy. In addition, the evaluation is conducted on a subset of the APPS benchmark, and while it provides a representative range of problem difficulties, it may not capture all characteristics of real-world coding tasks. Finally, both agents are instantiated from the same base model, which limits the diversity of perspectives compared to heterogeneous or ensemble-based approaches. These limitation translate to possible future work to be undertaken. One direction is to explore heterogeneous agent configurations, where the tutor and student are based on different models. Another direction is to develop more reliable stopping criteria, potentially by incorporating confidence estimation into the tutor’s decisions. Improving the quality of tutor feedback through additional training methods, such as preference-based DPO (Direct Preference Optimization), may also lead to further gains. More broadly, adapting the interaction structure based on problem difficulty could help allocate computational resources more effectively.

\section{CONCLUSION}
In this study, we proposed a developmental scaffolding framework PETITE inspired by peer tutoring interaction and instantiated it in the autonomous coding domain, a lightweight multi-agent system in which both tutor and student agents are derived from the same base LLM. Rather than relying on stronger supervisory models, distillation pipelines, or heterogeneous ensembles, the framework improves performance through structured internal interaction based on role differentiation.
A key aspect of our approach is its asymmetric, serial refinement mechanism combined with decision-based early stopping. By separating generative and evaluative processes into distinct roles, the framework leverages the model’s capacity for critical assessment in a way that is not typically engaged during solution generation alone. The tutor provides targeted, correctness-oriented feedback, while the student focuses on solution synthesis, enabling iterative improvement without external supervision or consensus across multiple agents.

Empirical results on the APPS benchmark show that PETITE achieves competitive success rates among iterative multi-agent approaches while using substantially fewer tokens than debate-based and multi-reviewer systems. These findings support the hypothesis that structuring interaction within a single model, through role separation, can yield performance gains while maintaining computational efficiency. More broadly, this work shows the benefit of grounding LLM-based problem-solving systems in principles from developmental science. Rather than treating multi-agent design as an engineering heuristic, our results suggest that such developmentally motivated interaction structures can serve as a principled basis for improving learning and problem solving under resource constraints. This perspective points toward integrating insights from human cognitive development into the design of more efficient and adaptable AI systems.

\addtolength{\textheight}{-12cm}   



\section*{ACKNOWLEDGMENT}
This work was supported by JSPS KAKENHI Grants numbered JP23K24926 and JP25H01236. We thank the developers of the APPS benchmark and the Qwen model team.  


\end{document}